\documentclass[preprint,12pt]{elsarticle}

\usepackage{amssymb}
\usepackage{amsmath}
\usepackage{mathptmx} 
\usepackage{amsmath} 
\usepackage{graphics} 
\usepackage{epsfig} 

\usepackage{times} 
\usepackage{amsmath} 
\usepackage{amssymb}  
\usepackage{graphicx} 
\usepackage{float} 
\usepackage{subfigure} 
\usepackage{parskip}
\usepackage[utf8]{inputenc}
\usepackage{float} 
\usepackage{amsmath}
\usepackage{amssymb}
\usepackage{booktabs}
\usepackage{graphicx} 
\usepackage{indentfirst} 
\usepackage{hyperref}
\usepackage{soul} 
\usepackage{xcolor} 
\sethlcolor{yellow} 
\renewcommand{\hl}[1]{#1}

\journal{}
\begin{document}

\begin{frontmatter}



\title{TSCnet: A Text-driven Semantic-level Controllable Framework for Customized Low-Light Image Enhancement}



\author{Miao Zhang\textsuperscript{a*}} 
\ead{zhangmiao@sz.tsinghua.edu.cn}
\affiliation{organization={Shenzhen International Graduate School, Tsinghua University}, 
            addressline={University Town of Shenzhen, Nanshan District}, 
            city={Shenzhen},
            postcode={518055}, 
            state={Guangdong},
            country={China}
            }
\affiliation{
    organization={Johns Hopkins University}, 
    addressline={3400 N. Charles Street}, 
    city={Baltimore},
    postcode={MD 21218}, 
    state={Maryland},
    country={USA}
}



\author{Jun Yin\textsuperscript{a*}} 
\ead{yinj24@mails.tsinghua.edu.cn}
\author{Pengyu Zeng\textsuperscript{a}} 
\ead{zeng-py24@mails.tsinghua.edu.cn}
\author{Yiqing Shen\textsuperscript{b}} 
\ead{yshen92@jhu.edu}
\author{Shuai Lu\textsuperscript{a$^{\dag}$}}
\ead{shuai.lu@sz.tsinghua.edu.cn}
\author{Xueqian Wang\textsuperscript{a}} 
\ead{wang.xq@sz.tsinghua.edu.cn}

\cortext[cor1]{\textsuperscript{*}These authors contributed equally to this work.}
\cortext[cor1]{\textsuperscript{\dag}Corresponding Author}


\begin{abstract}
\hl{Deep learning-based image enhancement methods show significant advantages in reducing noise and improving visibility in low-light conditions. These methods are typically based on one-to-one mapping, where the model learns a direct transformation from low light to specific enhanced images. Therefore, these methods are inflexible as they do not allow highly personalized mapping, even though an individual's lighting preferences are inherently personalized.
To overcome these limitations, we propose a new light enhancement task and a new framework that provides customized lighting control through prompt-driven, semantic-level, and quantitative brightness adjustments. The framework begins by leveraging a Large Language Model (LLM) to understand natural language prompts, enabling it to identify target objects for brightness adjustments. To localize these target objects, the Retinex-based Reasoning Segment (RRS) module generates precise target localization masks using reflection images. Subsequently, the Text-based Brightness Controllable (TBC) module adjusts brightness levels based on the generated illumination map. Finally, an Adaptive Contextual Compensation (ACC) module integrates multi-modal inputs and controls a conditional diffusion model to adjust the lighting, ensuring seamless and precise enhancements accurately.
Experimental results on benchmark datasets demonstrate our framework's superior performance at increasing visibility, maintaining natural color balance, and amplifying fine details without creating artifacts. Furthermore, its robust generalization capabilities enable complex semantic-level lighting adjustments in diverse open-world environments through natural language interactions.}\\
Project page is \href{}{https://miaorain.github.io/lowlight09.github.io/}
\end{abstract}




\end{frontmatter}


\section{Introduction}
\begin{figure}[htbp!]
    \centering
    \includegraphics[width=0.9\linewidth]{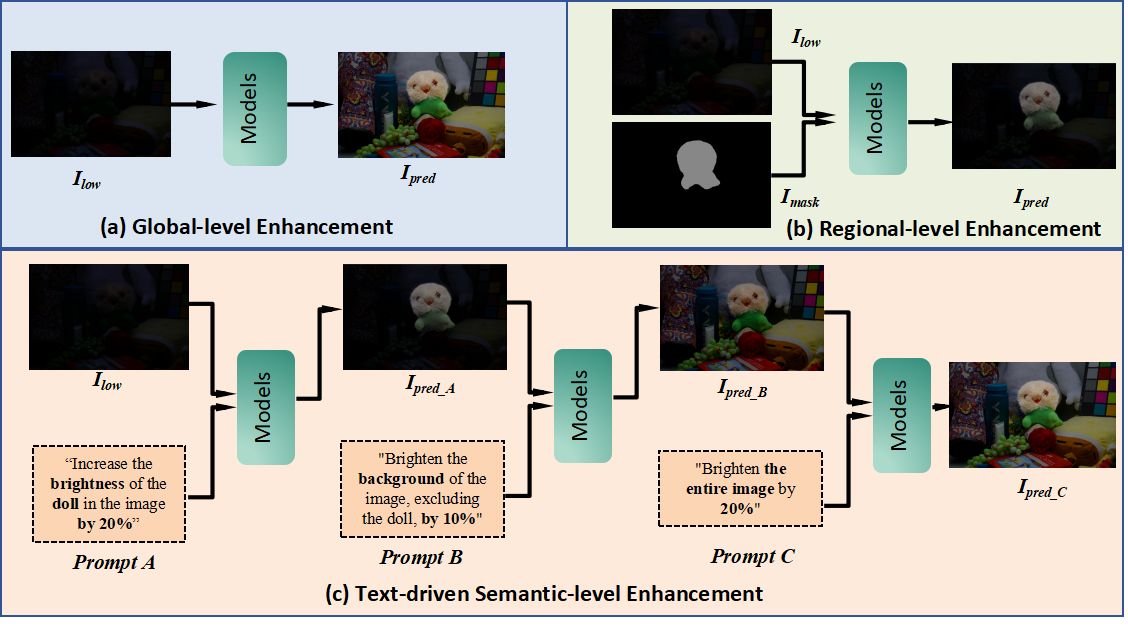}
    \caption{An overview of low-light enhancement tasks. The figure below illustrates three levels of low-light image enhancement tasks:
(a) Global-level Enhancement, which involves uniform brightness adjustment across the entire image;
(b) Regional-level enhancement, which includes selective enhancement of certain portions of an image, such as the background or the object of interest;
(c) Text-driven Semantic Enhancement, in which natural language prompts are used to adjust the brightness of targeted regions or the entire image, guided by semantic understandings.}
    \label{fig:intro}
\end{figure}

Many real-world applications require the use of low-light image enhancement (LLIE), including surveillance, traffic monitoring, and driving at night. The LLIE is capable of producing clear images even in dim lighting conditions, allowing users to identify potential hazards as well as important information more easily \cite{HLAFACE}. As illustrated in Figure \ref{fig:intro}, low-light images can be enhanced in two ways: global-level enhancement and regional-level enhancement, which can also be referred to as task a and task b, respectively.

Researchers have proposed various methods for global-level enhancement, including curve adjustment and histogram leveling, to optimize pixel distribution and produce sharper, more highlighted images \cite{HISTO-1, HISTO-2}. 
The problem with these methods is that they fail to capture the inherent pattern of pixel values, which causes color distortion and loss of detail in enhanced images.
\hl{With the development of deep learning (DL), a significant amount of DL models have been proposed and adapted to } \cite{ hai2023r2rnet, wang2024division,xu2024degraded, wang2023low, mythiliradial, yan2024dynamic, zhang2013multi, yan2024uncertainty}.  Based on the Retinex theory, Li et al. \cite{Dmph-net2023} presented LightenNet, a convolutional neural network designed to enhance low-light images. However, it is noted that LightenNet, may amplify noise in real scenes. Additionally, gamma correction is an effective technique to enhance pixel intensity in dark areas and adjust contrast in low-light images \cite{gamma_correction2022}. Although the DL as mentioned above methods can enhance brightness, they mainly focus on global brightness adjustments. Nonetheless, when lighting conditions are complex and light requirements vary from person to person, a greater degree of regional lighting adjustment is necessary.

To achieve regional-level enhancement, previous research started to incorporate semantic information into low-light enhancement tasks\cite{xu2022recoro, yin2023cle}. 
For the first time, ReCoRo~\cite{xu2022recoro} investigates this scenario by incorporating domain-specific augmentations into an enhancement model. While their enhancements are specifically designed for portrait images, their refinement may be required to accommodate masks of other image types. In addition, a conditional diffusion model is used by Yin /textit[et al.]\cite{yin2023cle} to enhance the brightness of specific regions of an image to any desired level through iterative refinement. 

Generally, global-level light enhancement methods employ a one-to-one mapping approach, limiting their flexibility for fine-tuning based on individual preferences. Nevertheless, regional-level light enhancement algorithms can adjust brightness at the semantic level, but this control still heavily depends on manually annotated masks \cite{xu2022recoro, yin2023cle}, limiting both flexibility and scalability. Moreover, people tend to prefer using natural language over manual clicking for light control. Therefore, we propose a new text-driven semantic-level light enhancement task, referred to as the c task in Figure \ref{fig:intro},  with the primary objective of enabling semantic-level, quantitative brightness adjustments through natural language control, allowing users to \textbf{"Adjust the Light as You Say"}.

The primary contributions of this work are outlined as follows.
\begin{itemize}
    \item 
    We first propose an new task, the Text-driven Semantic-level Light Enhancement task. Subsequently, we introduce a novel method, TSCnet, which utilizes a large language model to decompose language instructions into two distinct tasks: target localization and brightness enhancement.In order to accomplish this objective, we propose two modules. First, the Retinex-based Reasoning Segment (RRS) module generates mask images without manual annotation. Second, Text-based Brightness Controllable (TBC) module introduces illumination images to guide local lighting adjustments, ensuring the consistency of locally.
    \item 
    In order to integrate lighting information seamlessly with target objects, we also introduce the Adaptive Contextual Compensation (ACC) Module, which employs cross-attention mechanisms and channel adaptation to integrate lighting information seamlessly with target objects. Furthermore, ACC improves overall enhancement processes by efficiently coordinating additional data.
    \item 
     Comprehensive experiments were conducted on several benchmark datasets to demonstrate our method's superiority over existing state-of-the-arts. We showed our method's generalization capability and ability to perform personalized lighting adjustments using natural language. Additionally, to confirm the effectiveness of our proposed architecture, we performed an ablation study.
\end{itemize}

\section{Related Work}
\label{sec:review}
Deep learning techniques used for Low-Light Image Enhancement (LLIE) can be primarily divided into two categories: global-level enhancement \cite{Dmphnet2023He, llnet2023,shearlet_transform_2020, retinex-1, retinex-2, zhang2024retinex, wang2025mdanet, zhang2023scrnet} and regional-level enhancement \cite{xu2022recoro, yin2023cle, zhang2025tscnet}.  
\subsection{Global-level Low-light image enhancement}
In the realm of global-level enhancement, a number of architectures have emerged within LLIE which are specifically designed for low-light environments, such as noise reduction and detail preservation.
For instance, in the DMPH-Net architecture \cite{Dmphnet2023He}, a multi-scale pyramid structure is integrated with attention mechanisms to enhance low-light images while simultaneously minimizing noise.
In particular, EnlightenGAN \cite{jiang2021enlightengan} and LEGAN \cite{tao2024legan} stand out as models that utilize generative adversarial networks (GANs) to map low-light images to their improved versions. Due to their ability to handle typical low-light artifacts and noise, these architectures have shown exceptional performance in enhancing image quality. 
Furthermore, by incorporating targeted loss functions and innovative training methodologies, the performance of LLIE models has been improved. As an example, \cite{zhang2024retinex} introduced a deep learning technique inspired by the Retinex theory that improves images by learning how to map illumination onto itself, while integrating a superresolution process to improve output quality \cite{retinex2021liu}. 

\subsection{Regional-level Low-light image enhancement}
For regional-level low light enhancement \cite{xu2022recoro, yin2023cle}, ReCoRo \cite{xu2022recoro} integrates a control mechanism that allows users to define the region and desired illumination level to enhance the input image. However, the lack of objective metrics and paired annotations makes it difficult to obtain an aesthetically pleasing outcome from inaccurate user masks, which is a challenge the model aims to address but does not fully resolve. Besides, CLE Diffusion \cite{yin2023cle} introduces a new diffusion approach for controllable light enhancement, which allows for seamless brightness control during inference and the incorporation of the Segment-Anything Model (SAM) for easy region-specific enhancement with a single click.  However, this approach requires manual mask annotation, which is not user-friendly and does not allow for precise quantitative light adjustment.

\subsection{Diffusion model in low light enhance}

In diffusion models, data distributions are learned by gradually adding noise to the data and training neural networks to reverse the process, thereby effectively denoising the data. As a result, it is highly effective for enhancing low-light images due to its ability to capture complex patterns and structures. 
For example, Nguyen et al. \cite{nguyen2024diffusion}   apply a cascaded technique with Iterative Latent Variable Refinement (ILVR) to achieve a high level of reconstruction quality and consistency even in low-light conditions. The study by Jiang et al \cite{jiang2023low} incorporates a high-frequency restoration module (HFRM) to enhance fine-grained detail reconstruction by complementing diagonal features with vertical and horizontal data. In addition, Shang et al \cite{shang2024multi} have introduced both spatial and frequency domain information to enhance low light imaging, which allows for more precise and detailed information to be captured. In \cite{hou2024global}, a global structure-aware regularization strategy for diffusion-based frameworks is presented. This strategy promotes structural and content coherence across similar regions of enhanced images, improving their naturalness and aesthetic quality while maintaining fine details and textures. Besides, according to Wang et al. \cite{wang2023exposurediffusion}, a dynamic residual layer adjusting denoising approach according to the noise-to-signal ratio is presented, which effectively reduces side effects during iterative enhancement when intermediate outputs are sufficiently exposed. Overall, the optimization of diffusion architecture and the integration of additional information are essential for improving low-light enhancement.

\section{The Proposed Method}

\subsection{Problem Formulation and Overall Architecture}
\label{sec:problem}
Light is intricately linked to human life \cite{ma2025street, li2024gagent, zhang2024adagent}. In practice, the requirements for lighting adjustments vary greatly from person to person. Furthermore, natural language is typically favored over manual annotations \cite{yin2023cle} for controlling semantic-level vision tasks due to its intuitive and user-friendly nature. However, natural language is inherently ambiguous and abstract \cite{he2025enhancing1, he2025enhancing2}, which complicates its translation into precise actions. Consequently, two key challenges must be addressed to achieve text-driven semantic-level light controllable enhancement:

\begin{itemize}
    \item \textbf{How natural language instructions may be effectively converted into executable and customized quantitative conditions.}
    \item \textbf{How to implement customized lighting control based on the conditions outlined above.}
\end{itemize}
To address these challenges, we innovatively present a Large Language Model (LLM) to interpret abstract and vague concepts within natural language commands, accurately identifying target objects for enhancement and quantifiable brightness levels. Then, the Retinex-based Reasoning Segment (RRS) module is utilized to achieve semantic-level target localization (\( I_{\text{mask}} \)), followed by the Text-based Brightness Controllable (TBC) module to estimate the quantitative lighting adjustments (\(M_{\text{adjustment}}\)). Finally, the Adaptive Contextual Compensation (ACC) module adaptively integrates (\( I_{\text{mask}} \)) and (\(M_{\text{adjustment}}\)) as additional customized conditions into a control diffusion model to achieve context-aware, semantic-level diffusion-based light enhancement.

\begin{figure*}[htbp]
    \centering
    \includegraphics[width=1\linewidth]{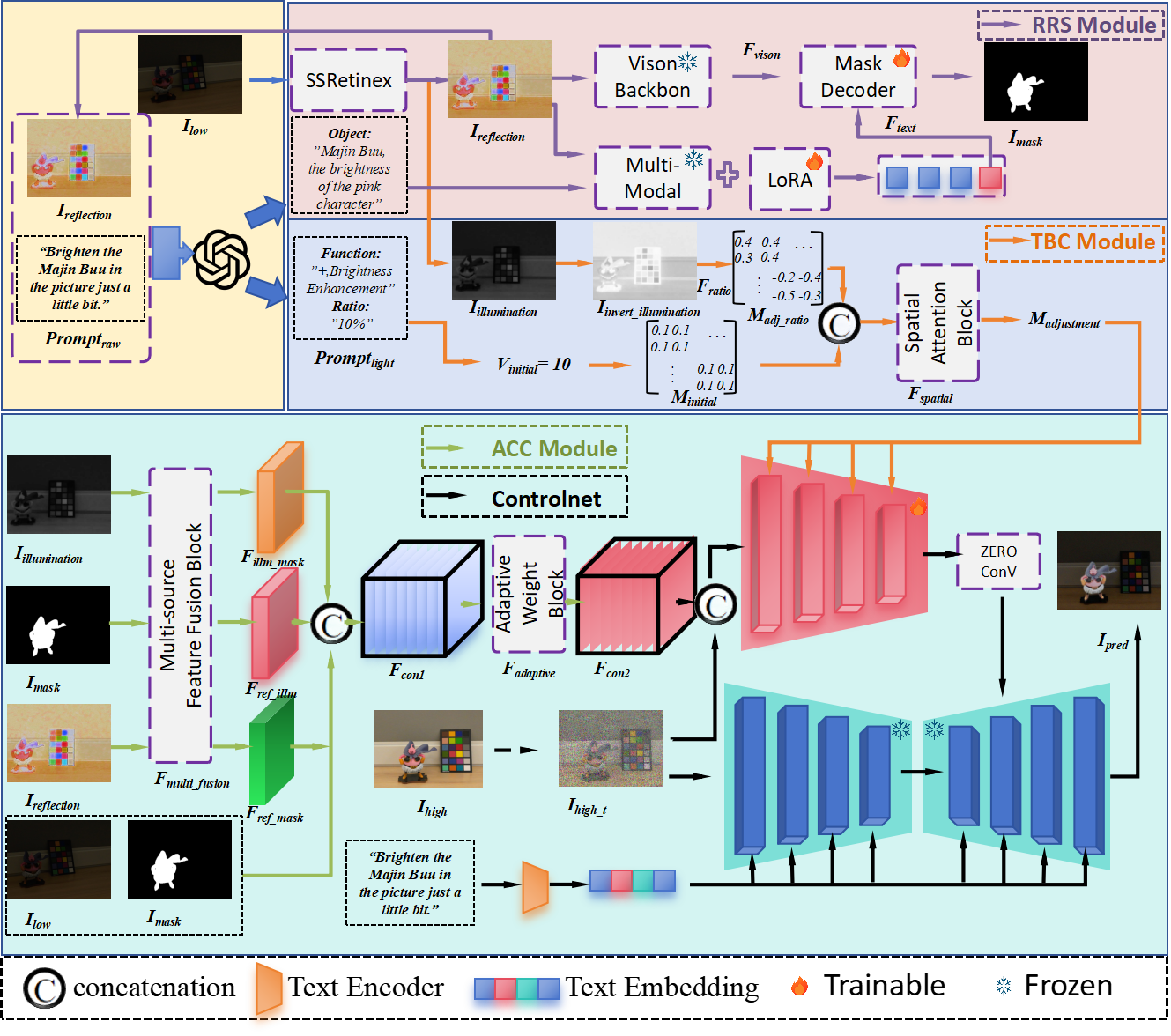}
    \caption{The overview of our framework, including the RRS Module, TBC Module, ACC Module and control diffusion. The framework begins with input prompts processed through a text encoder, generating task-specific adjustments such as brightness modification. The Retinex-based Reasoning Segment (RRS) identifies target regions via multi-modal input, enhanced by a LoRA-based mechanism. The Text-based Brightness Controllable (TBC) module applies precise brightness modifications using spatial attention. Finally, the Adaptive Contextual Compensation (ACC) module integrates multi-source inputs, ensuring coherent image enhancement through adaptive weight fusion and ControlNet, driving the conditional diffusion process to achieve high-quality, personalized low-light image enhancement.}
    \label{fig:stage1}
    \vspace{-0.25cm}
\end{figure*}

\subsection{Stage one: Target Object  Localization and Quantitative Brightness Estimation}

\noindent\textbf{Preprocessing Workflow.} 
Preprocessing begins with the Large Language Model (LLM), specifically GPT-4o, which processes natural language input  \( \text{Prompt}_{\text{raw}} \) such as "Brighten the Majin Buu in this picture just a little." LLM recognizes both target object (e.g. Majin Buu; the pink character) and necessary action (10\% brightening).

\noindent\textbf{The Retinex-based Reasoning Segment (RRS) Module.} 
To achieve semantic-level light adjustment, obtaining accurate semantic information is important. Existing research indicates that both methods ReCoRo \cite{xu2022recoro} and CLE \cite{yin2023cle} depend on human annotation, which is not only time-intensive and resource-demanding but also limits effective interaction with the task. In response to this challenge, inspired by \cite{zhang2024retinex, yan2024visa, lai2024lisa,qi2024generalizable}, the present study employs a reasoning segment, which directly identifies and acquires the precise image of the target object using natural language processing techniques. Furthermore, given that images captured under low-light conditions often lose detail \cite{bilateral, DEANet,cui2022IAT}, resulting in coarse segmentation, this study utilizes the decomposition network proposed by Retinex theory \cite{land1971lightness} to mask reflection images that contain rich texture information \cite{Chen2018Retinex, retinex-1,retinex-3,retinex-d3}, thereby producing more refined and accurate images of the target object.The specific steps are as follows:

The Retinex-based Reasoning Segment (RRS) Module  achieves precise target localization and enhancement by decomposing \( I_{\text{low}} \) as:

\[
I_{\text{low}} = I_{\text{illumination}} \times I_{\text{reflection}}
\]

Parsed natural language descriptions (e.g. "Majin Buu, the brightness of his pink character") are then processed through a Multi-Modal block known as \( F_{\text{text}} \) in \cite{lai2024lisa}, and this fuses textual features with visual features extracted from \( I_{\text{reflection}} \) to form visual features extracted by Vision Backbone: see \cite{yan2024visa} for further discussion on this process.

\begin{align}
 \vspace{-0.25cm}
F_{\text{fusion}} = \text{concat}(F_{\text{vision}}, F_{\text{text}})
\end{align}

This fusion of visual and textual features is then fed into the \textbf{Mask Decoder}\cite{lai2024lisa}, which predicts the segmentation mask \( I_{\text{mask}} \):
\vspace{-0.25cm}
\begin{align}
I_{\text{mask}} = F_{\text{decoder}}(F_{\text{fusion}})
\end{align}

This process ensures accurate mask generation for pixel-level segmentation, thus enabling detailed low-light enhancement based on natural language input.

\noindent\textbf{Text-based Brightness Controllable Module (TBC).} 
Considering the diversity of individual lighting needs, precise control of light quantity is important. Previous research methods have tackled this challenge in different ways: ReCoRo \cite{xu2022recoro} supports only binary brightness adjustment, Zero-DCE \cite{ZeroDCE} achieves it through different continuous iterations to optimize enhancement curves, and CLE \cite{yin2023cle} requires a specific numerical value to be specified for brightness regulation. Essentially, when people express their needs for personalized lighting, they often cannot provide precise quantitative parameters and tend to use vague language, such as "increase the brightness a bit". However, these methods are difficult to meet such requirements. Therefore, we use language models to parse specific quantitative parameters from vague instructions, and utilize the illumination map derived from the Retinex theory \cite{retinex-d4, retinex-d5} as a guide map, to achieve language-driven, more precise, region-consistent adjustment of lighting to meet personalized needs better.The detailed content is as follows.

The TBC Module handles the brightness control based on natural language input \textit{\( \text{Prompt}_{\text{bright}} \)}  as baseline and \( I_{\text{illumination}} \) as basis for varying brightness levels, serving as a spatial consistency guide map. The specific implementation is as follows.
\( I_{\text{illumination}} \) : Invert, Clip, Calculate the Mean, and Normalize. By leveraging the spatial attention mechanism \cite{woo2018cbam}, we derive:
\begin{align}
M_{\text{adjustment}} = 
F_{\text{spatial}}(M_{\text{initial}}, M_{\text{adj\_ratio}})
\end{align}
In all, the structure of the TBC Module enables complex, localized brightness adjustments guided by illumination maps, resulting in consistent outcomes that are closely aligned with the requirements of the user.

\subsection{Stage Two: Adaptive Contextual Compensation and Controllable Denoising Process}
\noindent\textbf{Adaptive  Contextual Compensation (ACC) Module.} 

It is important to incorporate additional information into diffusion models \cite{zhang2023adding, ye2023ip, mou2024t2i, ruiz2023dreambooth, huang2023composer} in order to enhance task performance. In contrast to previous methods \cite{mou2024t2i} and \cite{rombach2022high} that merely add or concatenate multiple sources of information, this study incorporates cross-attention modules \cite{huang2019ccnet} and global channel modules \cite{woo2018cbam} that facilitate adaptive fusion of multimodal resources. In this module, the model is enhanced to be capable of effectively interpreting diverse input data and optimizing weight distribution, which enables the model to generate more precise lighting dynamics and enhance image brightness. 

The three key inputs, including \( F_{\text{illumination}} \), \( F_{\text{mask}} \), and \( F_{\text{reflection}} \), are first processed through spatial depth-wise convolution layers to extract spatially and depth-wise enhanced features for each input. Subsequently, cross-attention mechanisms are applied to enhance the relationships between inputs, improve contextual understanding, and use other inputs as complementary resources to compensate for the deficiencies of a single input. Following the cross-attention process, the enhanced feature maps are combined with \(I_{\text{low}}\) and \(I_{\text{mask}}\) through channel-wise concatenation to form the refined feature maps (\(F_{\text{con1}}\)), which are then fed into global channel attention as adaptive guidance maps to dynamically adjust the weights of each feature map. Finally, this process generates the resulting feature maps (\(F_{\text{con2}}\)), seeing Figure \ref{fig:acc}:
\begin{figure*}[htp]
    \centering
    \includegraphics[scale=0.6]{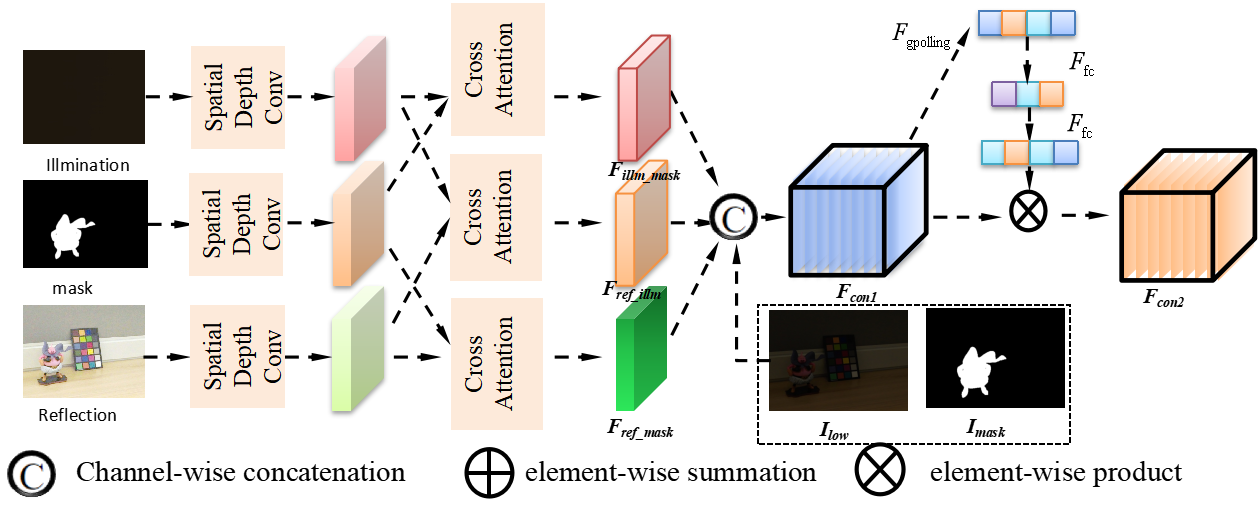}
    \caption{ The architecture of the Adaptive Contextual Compensation (ACC) Module. The module processes three key inputs: illumination, mask, and reflection, each passed through spatial depth-wise convolution layers. Cross-attention mechanisms are used to enhance the interaction of features between the  different inputs. In the following steps, channel-wise concatenation is followed by an element-wise summation operation to combine the resulting feature maps, followed by an element-wise product operation to refine and adapt the features. The final output feature maps \(F_{con1}\) and \(F_{con2}\) guide the lighting adjustments in the low-light image \(I_{low}\), ensuring context-aware, coherent image enhancement.
    }
    \label{fig:acc}
\end{figure*}
\begin{align}
F_{\text{illm\_mask}}, F_{\text{ref\_illm}}, F_{\text{ref\_mask}}  = F_{\text{multi\_fusion}}(F_{\text{illum}}, F_{\text{mask}}, I_{\text{reflection}})
\end{align}

\begin{align}
F_{\text{con2}} = F_{\text{adaptive}}(F_{\text{illm\_mask}}, F_{\text{ref\_illm}}, F_{\text{ref\_mask}},I_{\text{low}}, I_{\text{mask}})
\end{align}
This combined feature map, which integrates $Prompt_{raw}$ and $M_{adjustment}$, is then processed by a series of convolutional layers including adaptive cross attention\cite{huang2019ccnet,chen2021crossvit,huo2022three} and controlnet\cite{zhang2023adding,mou2024t2i}, leading to the final prediction of the high-quality image \( I_{\text{pred}} \):
\begin{align}
I_{\text{pred}} &= F(\boldsymbol{x} ; \Theta)+Z\left(F\left(\boldsymbol{x}+Z\left(c ; \Theta_{\mathrm{z} 1}\right) ; \Theta_{\mathrm{c}}\right) ; \Theta_{\mathrm{z} 2}\right) \\
c &= (F_{\text{con2}}, M_{\text{adjustment}}, Prompt_{\text{raw}})
\end{align}

\noindent\textbf{Controllable Denoising Module.} 

Diffusion models \cite{sohl2015deep} are generative models that progressively reverse a multi-step noise addition process, typically modeled as a Markov chain, to reconstruct data from random noise. As an example, consider the DDPM model \cite{ho2020denoising}, which performs both a forward and a reverse process, in which Gaussian noise is incrementally inserted into a clean image in the forward pass. This is mathematically represented as:
\begin{align}
q(y_t | y_{t-1}) = \mathcal{N}(y_t; \sqrt{1-\beta_t} y_{t-1}, \beta_t\mathbf{I})
\end{align}

However, this noise addition calculation process needs to be calculated step by step, and the calculation efficiency is relatively poor. At this time, a calculation technique of conditional Gaussian distribution can be used to directly calculate $y_t$ at any time from $y_0$ in one step.

\begin{equation}
\begin{aligned}
y_t\sim \mathcal{N}\left(\sqrt{\bar{\alpha}_t} y_0,\left(1-\bar{\alpha}_t\right) I\right), \bar{\alpha}_t=\prod_{i=1}^t 1-\beta_t
\end{aligned}
\end{equation}

To speed up the sampling process, DDIM \cite{song2020denoising} introduces a deterministic method, as described below:

\begin{equation}
\begin{aligned}
y_{t - 1} &= \sqrt{\alpha_{t - 1}}\left(\frac{y_t - \sqrt{1 - \alpha_t} \epsilon_\theta\left(y_t, t\right)}{\sqrt{\alpha_t}}\right) + \\
&\quad \sqrt{1-\alpha_{t - 1}-\sigma_t^2} \cdot \epsilon_\theta\left(y_t, t\right)+\sigma_t \epsilon_t
\end{aligned}
\end{equation}

Where, $\sigma_t^2=\eta \cdot \beta_t$. Here, \( \epsilon_\theta \), typically implemented using a U-Net architecture \cite{ronneberger2015u}, estimates the noise in a noisy image. Inference begins by sampling from \( y_T \sim \mathcal{N}(0, \mathbf{I}) \) until eventually returning back to its "clean image of \( y_0 \).

Controlnet is used as the core model to produce images that meet our requirements, typically being trained through minimizing its negative log-likelihood loss function (see the simplified form here):

\begin{equation}
\mathcal{L}_\text{simple}=\mathbb{E}_{\mathbf{y}_{0}, t, c,\epsilon}\left[\left\|\epsilon-\epsilon_{\theta}\left(\sqrt{\bar{a}_{t}} \mathbf{y}_{0}+\sqrt{1-\bar{a}_{t}} \epsilon,t,c\right)\right\|^{2}\right]
\end{equation}

\subsection{Auxiliary Loss Functions} 
To enhance the sensitivity of our generative model, we incorporate an supplementary loss\cite{yin2023cle} to directly supervise denoising estimation. Formally, the auxiliary loss function can be formulated as:

\begin{equation}
\begin{aligned}
\mathcal{L}_{\text{aux}} = &\ \mathcal{L}_{\text{base}} + W_{\text{col}} \mathcal{L}_{\text{col}} 
& + W_{\text{ssim}} \mathcal{L}_{\text{ssim}}
\end{aligned}
\end{equation}

where  \( \mathcal{L}_{\text{col}} \) is the Angular Color Loss\cite{wang2019underexposed}, and \( \mathcal{L}_{\text{ssim}} \) is the SSIM Loss\cite{hore2010image}, \( W_{\text{col}}, W_{\text{ssim}}\) are the weighting factors.

The color loss can be expressed as:
\begin{equation}
\mathcal{L}_\text{col}=\sum_{i}{\angle\left(\hat{y_0}_i,y_i\right)},
\end{equation} 
where $i$ denotes a pixel location and $\angle(,)$ determines the angle difference between two 3-dimensional vectors representing colors in RGB color space.

SSIM loss can be expressed as follows,
\begin{equation}
\mathcal{L}_\text{ssim} = \frac{(2\mu_{y}\mu_{\hat{y_0}} + c_1)(2\sigma_{y\hat{y_0}} + c_2)}{(\mu_{y^2} + \mu_{\hat{y_0}}^2 + c_1)(\sigma_{y^2} + \sigma_{\hat{y_0}}^2 + c_2)},
\end{equation} 
where $\mu_y$ and $\mu_{\hat{y_0}}$ are pixel value averages, $\sigma_y$ and $\sigma_{\hat{y_0}}$ are variances, $\sigma_{y\hat{y_0}}$ is covariance, $c_1$ and $c_2$ are constants for numerical stability.

\section{Experimental Results}

\subsection{Experimental Setup}
The experiment proceeded in two stages: In the first RRS module stage,  we fine-tuned a pre-trained U-Net using the AdamW optimizer, with a learning rate of 0.0003 no weight decay, and a batch size of 6.
In the second control net stage, the Adam optimizer was applied (initial learning rate of \(1 \times 10^{-4}\), decaying to \(1 \times 10^{-6}\) via cosine annealing), a batch size of 8, and weight decay of \(1 \times 10^{-5}\), over 1000 epochs.

\subsection{Datasets and Metrics}
We evaluate our model on two commonly employed benchmarks, LOL~\cite{wei2018deep} and MIT-Adobe FiveK~\cite{fivek}. The LOL dataset comprises 485 paired images for training and 15 paired images for testing, where every pair includes a low-light and a corresponding standard-light image. In the first RRS module, we generated 420 masked data pairs for training based on the LoL dataset. In the second ControlNet, to perform a quantitative task, we first use the Retinex algorithm \cite{kind_plus} to decompose the low-light image into an illumination (\( Low_{\text{illumination}} \)) image and a reflection (\( Low_{\text{reflection}} \)) image. The same process is applied to the high-light image, yielding \( High_{\text{illumination}} \) and  \( High_{\text{reflection}} \). Next, we adjust the illumination \( Low_{\text{illumination}} \) across ten levels and then combine it with the \( High_{\text{reflection}} \) to generate the ground truth image.
The MIT-Adobe FiveK dataset contains 5000 images edited by five experts using Adobe Lightroom. Following previous works~\cite{tu2022maxim,ni2020towards}, we use 4500 paired images for training and reserve 500 images for testing.
To assess the quality of the output images, we use SSIM~\cite{wang2004image}, LI-LPIPS~\cite{zhang2018unreasonable}, LPIPS~\cite{zhang2018unreasonable}, and PSNR~\cite{huynh2008scope} metrics.

\subsection{Comparison Experiment}
\begin{figure*}[htbp]
    \centering
    \includegraphics[width=0.9\linewidth]{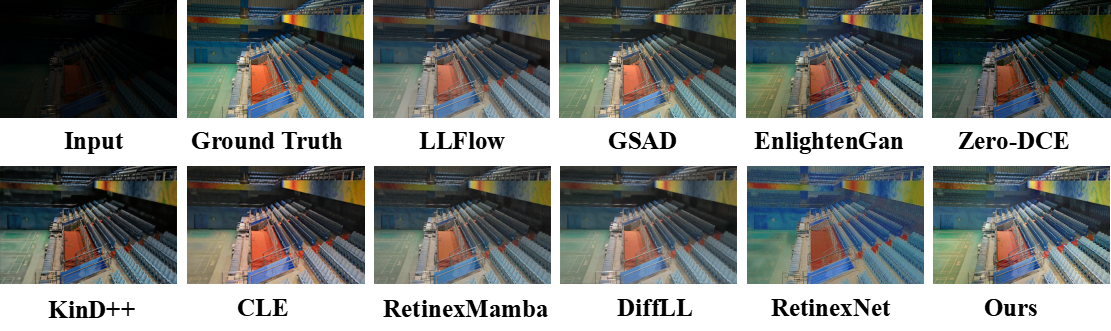}
    \caption{Visual comparison with other advanced approaches on LOL.}
    \label{fig:comparison_lol}
\end{figure*}

\begin{figure*}[htbp]
    \centering    
    \includegraphics[width=0.9\linewidth]{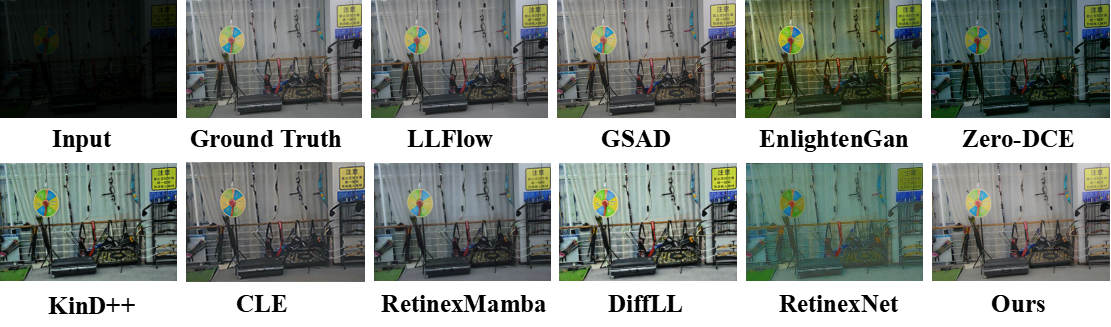}
    \caption{Visual comparison with other advanced approaches on LOL2.}
    \label{fig:comparison_lol2}
\end{figure*}

\begin{figure*}[htbp]
    \centering    
    \includegraphics[width=0.9\linewidth]{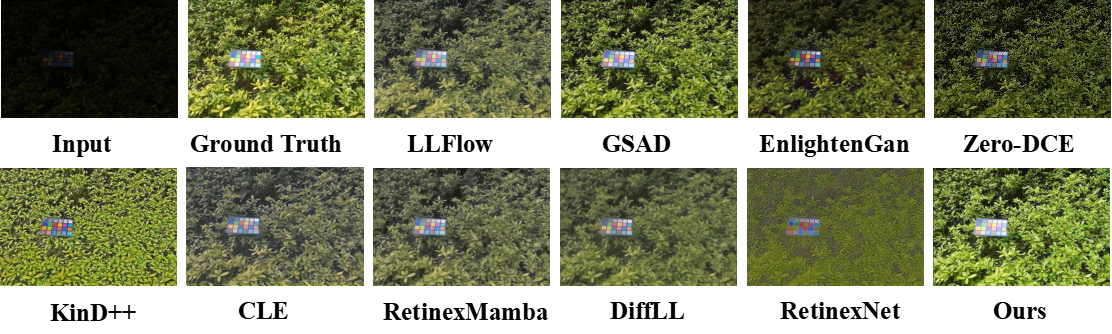}
    \caption{Visual comparison with other advanced approaches on LOM.}
    \label{fig:comparison_LOM}
\end{figure*}

\begin{figure*}[htbp]
    \centering    
    \includegraphics[width=0.9\linewidth]{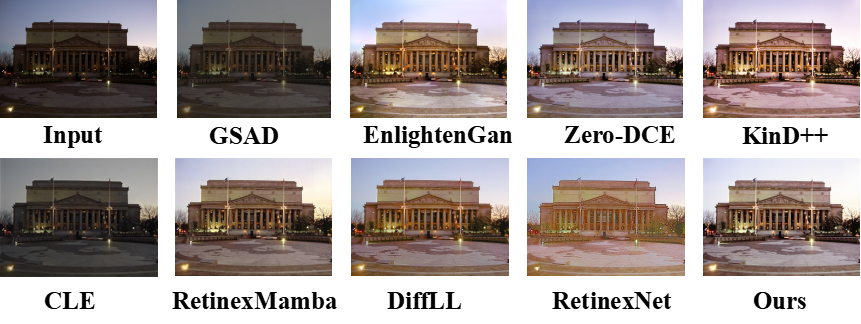}
    \caption{Visual comparison with other advanced approaches on DICM.}
    \label{fig:comparison_DICM}
\end{figure*}

\begin{figure*}[htbp]
    \centering    
    \includegraphics[width=0.9\linewidth]{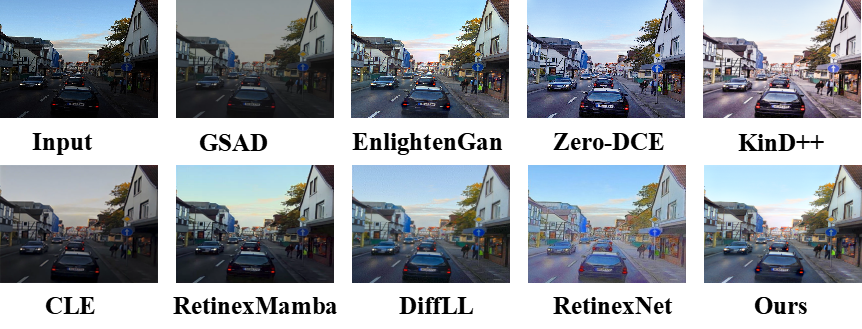}
    \caption{Visual comparison with other advanced approaches on LIME.}
    \label{fig:comparison_LIME}
\end{figure*}

\begin{figure*}[htbp]
    \centering    
    \includegraphics[width=0.9\linewidth]{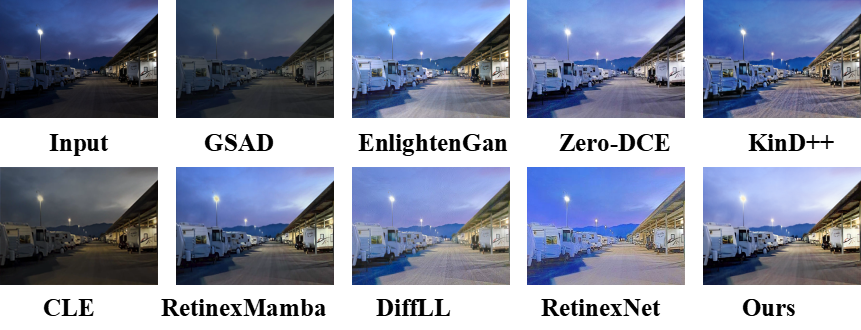}
    \caption{Visual comparison with other advanced approaches on NPE.}
    \label{fig:comparison_NPE}
\end{figure*}

\begin{figure*}[htbp]
    \centering    
    \includegraphics[width=0.9\linewidth]{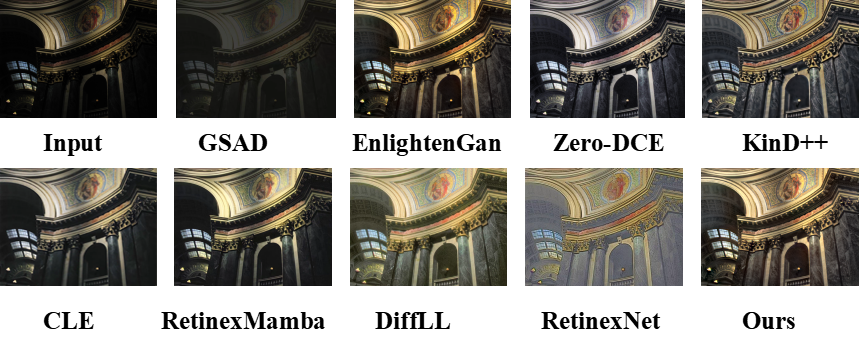}
    \caption{Visual comparison with other advanced approaches on MEF.}
    \label{fig:comparison_MEF}
\end{figure*}

\begin{figure*}[htbp]
    \centering    
    \includegraphics[width=0.9\linewidth]{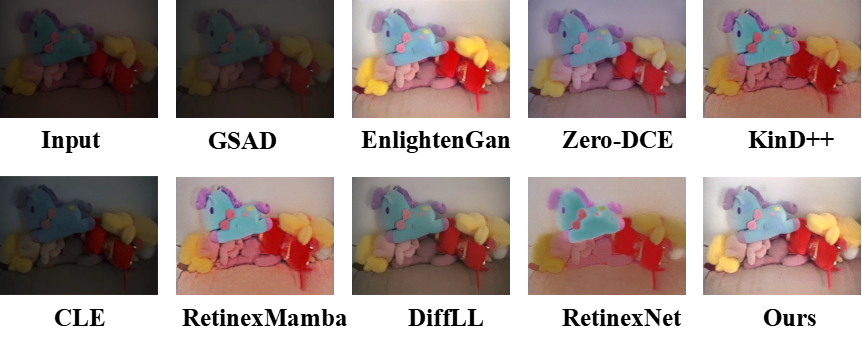}
    \caption{Visual comparison with other advanced approaches on REAL.}
    \label{fig:comparison_REAL}
\end{figure*}

\noindent\textbf{Qualitative Comparison.}
Figures from \ref{fig:comparison_lol} to \ref{fig:comparison_REAL} demonstrate the effectiveness of various methods, including LLFlow~\cite{wang2022low},  GSAD~\cite{hou2024global}, EnlightenGAN~\cite{jiang2021enlightengan}, Zero-DCE~\cite{Zero-DCE}, KinD++~\cite{zhang2021beyond}, CLE~\cite{yin2023cle}, RetinexMamba~\cite{bai2024retinexmamba},
DiffLL~\cite{jiang2023low},
Retinex-Net~\cite{wei2018deep},
and Our method. We have demonstrated the effectiveness of our method in improving visibility in low light conditions. It has been shown that our method provides marked improvements in contrast, brightness, detail preservation without creating unnatural artifacts, whereas other methods, such as GSAD~\cite{hou2024global}  and EnlightenGAN~\cite{jiang2021enlightengan}, either overexpose or under-enhance certain areas, especially complex lighting scenarios such as DICM~\cite{lee2013contrast} and LIME~\cite{guo2016lime}.

\begin{figure*}
\centering
\includegraphics[width=0.9\textwidth,keepaspectratio]{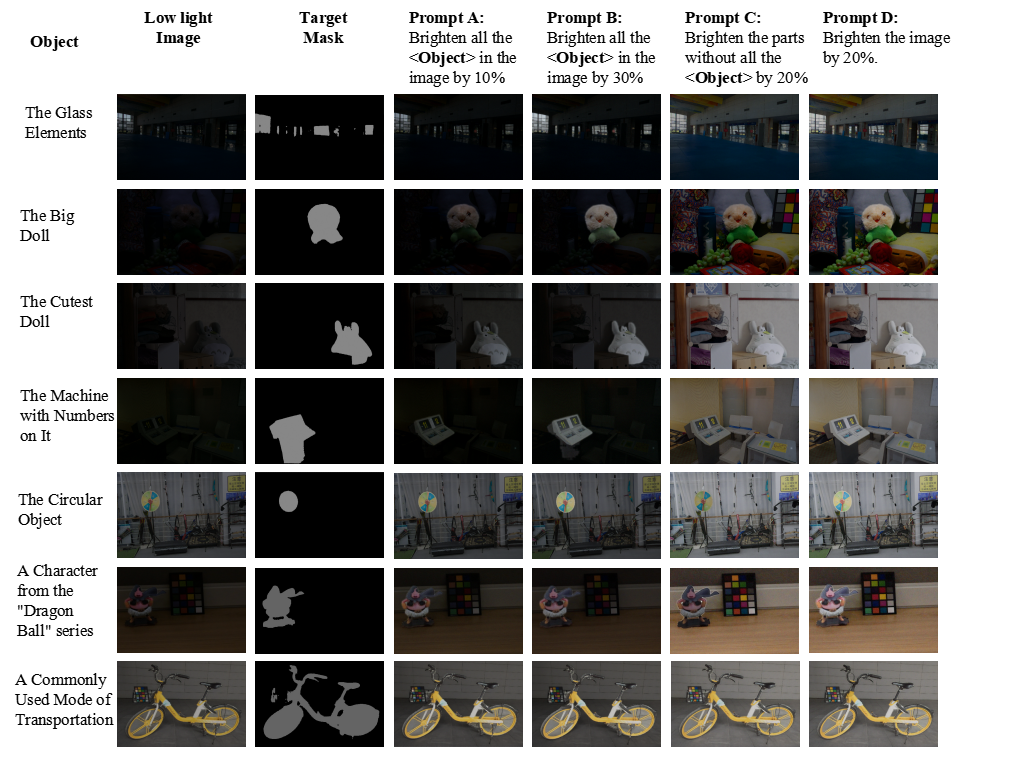}
\caption{
 A comparison of low-light image processing using different prompts for object-specific and general brightness adjustments. In each task, natural language instructions are used to brighten either target or background lighting, all driven by natural language prompts. }
\vspace{-3mm}
\label{fig:flexible_adjustment}
\end{figure*}
Figure ~\ref{fig:flexible_adjustment} demonstrates how natural language can be successfully applied to control complex lighting adjustments in images with both precision and flexibility. According to prompt input, the framework adjusts the brightness of objects, such as dolls, circles, or machines, according to a specified percentage. Furthermore, our framework not only changes the lighting of the target objects, but also changes the lighting conditions of the background, or the whole image. These samples clearly demonstrate our framework has successfully achieved text-driven semantic-level light control.

\noindent\textbf{Quantitative Comparison.} 
Table ~\ref{tab:lol} provides a comparison of state-of-the-art methods tested on the LOL dataset\cite{wei2018deep}. The table provides a comparison between each approach using key metrics such as PSNR (Peak Signal-to-Noise Ratio)\cite{huynh2008scope}, SSIM (Structural Similarity Index Measure)\cite{wang2004image}, LPIPS (Learned Perceptual Image Patch Similarity) \cite{zhang2018unreasonable} and LI-LPIPS (Light Independent LPIPS) \cite{yin2023cle}. Our approach stands out by achieving the highest PSNR score of 25.78 among all competing models, such as LLFlow (25.19) and CLE (25.51). SSIM scores for our model were excellent at 0.94, second only to MAXIM's 0.96, and showed strong structural preservation in enhanced images. Our model and LLFlow achieved similar results for LPIPS, which measures perceptual similarity, while both achieved similar perceptual quality scores of 0.15.

\begin{table*}
\centering
\caption{Comparisons on the LOL dataset.The best result is indicated in \textbf{bold}, while the second-best result is underlined.}
\begin{tabular}{lcccc}
\toprule 
Method                 & PSNR↑ & SSIM↑ & LPIPS↓ & LI-LPIPS↓ \\
\midrule
Zero-DCE~\cite{guo2020zero}    & 14.86 & 0.54  & 0.33 & 0.3051   \\
EnlightenGAN~\cite{jiang2021enlightengan} & 17.48 & 0.65  & 0.32  &0.2838  \\
RetinexNet~\cite{Chen2018Retinex}  & 16.77 & 0.56  & 0.47 &0.5468   \\
DRBN~\cite{yang2020fidelity}    & 20.13 & 0.83  & 0.16 &0.3271   \\
KinD++~\cite{zhang2021beyond}   & 21.30  & 0.82  & 0.16  &0.3768  \\
MAXIM~\cite{tu2022maxim}  & 23.43 & \textbf{0.96}  & 0.20  &\underline{0.1801}      \\
HWMNet ~\cite{fan2022half}     &  24.24 & 0.85  & 0.12  &0.1893  \\
LLFlow ~\cite{wang2022low} &25.19  &0.93  &\textbf{0.11}  & \textbf{0.1763}    \\

SCRnet ~\cite{zhang2024retinex} &23.16  &0.84  &0.21  & 0.1902    \\

CLE ~\cite{yin2023cle}      & \underline{25.51} &0.89  & 0.16   & 0.1841 \\
\textbf{Ours}        & \textbf{25.78} &  \underline{0.94}  & \underline{0.15}   & 0.1814  \\
\bottomrule
\end{tabular}
\label{tab:lol}
\end{table*}

The table ~\ref{tab:mit} presents a comparison of several advanced approaches on the MIT-Adobe FiveK dataset, evaluating their performance using two key metrics: PSNR and SSIM. TSCnet demonstrates the best performance with a PSNR of 29.92, surpassing CLE, which has the second-best result at 29.81. 
Other notable performances include HWMNet (26.29) and MAXIM (26.15).CLE attains the highest SSIM value of 0.97, while our method and HWMNet share the second-best score of 0.96. This demonstrates that our method performs exceptionally well in maintaining structural similarity.

\begin{table}[htbp]
\centering
\caption{Comparisons on the MIT-Adobe FiveK dataset. The best result is indicated in \textbf{bold}, while the second-best result is underlined.}
\setlength{\tabcolsep}{3mm}
\begin{tabular}{lcc}
\toprule 
Method       & PSNR↑ & SSIM↑   \\
\midrule
EnlightenGAN~\cite{jiang2021enlightengan} & 17.74 & 0.83    \\
CycleGAN ~\cite{zhu2017unpaired}  & 18.23 & 0.84    \\
Exposure~\cite{hu2018exposure} & 22.35 & 0.86    \\
DPE ~\cite{chen2018deep}        & 24.08 & 0.92    \\
UEGAN ~\cite{ni2020towards}        &  25.00 & 0.93  \\
MAXIM ~\cite{tu2022maxim}      & 26.15 & 0.95  \\
HWMNet ~\cite{fan2022half}        & 26.29 & 0.96     \\
SCRnet ~\cite{zhang2024retinex}  &  26.16 &0.92 \\
CLE ~\cite{yin2023cle}         &  \underline{29.81} &\textbf{0.97}   \\
\textbf{Ours}        &  \textbf{29.92} &  \underline{0.96} \\
\bottomrule
\end{tabular}
\label{tab:mit}
\vspace{-3mm}
\end{table}

We also perform experiments on unpaired real-world low-light images, including DICM~\cite{lee2013contrast} , LIME~\cite{guo2016lime}, MEF~\cite{ma2015perceptual}, NPE, and VV, using the NIQE metric for image quality assessment.Table ~\ref{tab:unpaired}  shows the quantitative results of our evaluations, where our method consistently achieves competitive NIQE scores across various datasets. For DICM, our model performs best with a score of 3.68, and on LIME, we achieve 4.14, second only to DiffLL. 
\begin{table}[ht]
\centering
\caption{EVALUATION RESULTS IN TERMS OF NIQE ON DICM, LIME, MEF,
NPE, AND VV DATASETS, where the best result is indicated in \textbf{bold}, while the second-best result is underlined.}
\resizebox{0.6\linewidth}{!}{
\begin{tabular}{lccccc}
\hline
\textbf{Method} & \textbf{DICM} & \textbf{LIME} & \textbf{MEF} & \textbf{NPE} & \textbf{VV} \\
\hline
RetinexNet  & 4.33 & 5.75 & 4.93 & 4.95 & 4.32 \\
KinD        & 3.95 & 4.42 & 4.45 & 3.92 & 3.72 \\
Zero-DCE    & 4.58 & 5.82 & 4.93 & 4.53 & 4.81 \\
EnlightenGAN& 4.06 & 4.59 & 4.70 & 3.99 & 4.04 \\
Zero-DCE++  & 4.89 & 5.66 & 5.10 & 4.74 & 5.10 \\
KinD++      & 3.89 & 4.90 & 4.55 & 3.91 & 3.82 \\
DCC-Net     & \underline{3.70} & 4.42 & 4.59 & 3.70 & \textbf{3.28} \\
SNR-Aware	& 6.12 & 5.93 & 6.44 & 6.44 & 11.5 \\
DiffLL	    & 3.87 & \textbf{3.78} & \textbf{3.43}	 & \textbf{3.43}	& \underline{3.51} \\
\textbf{Ours}	    & \textbf{3.68} & \underline{4.14} & \underline{3.92} & \underline{3.66}	& 3.70 \\
\hline
\label{tab:unpaired}
\end{tabular}
}
\end{table}

\subsection{Ablation Study}
\noindent\textbf{Ablation of Module Architecture.} 
\begin{figure}
\centering
\includegraphics[width=0.9\textwidth,keepaspectratio]{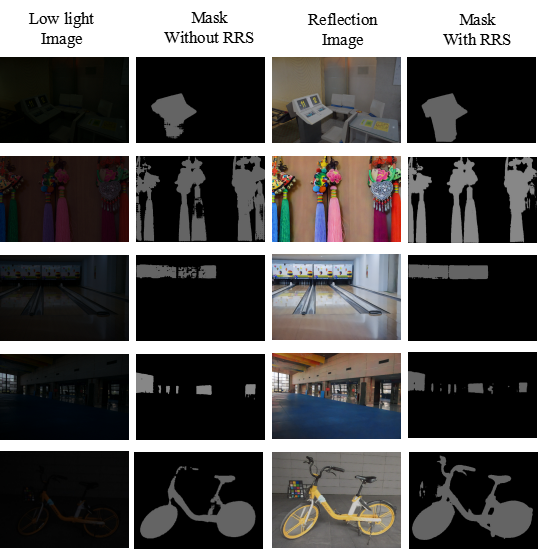}
\vspace{-4mm}
\caption{Comparison of mask generation without and with RRS module.}
\vspace{-8mm}
\label{fig:mask_ablation_study}
\end{figure}

Figure ~\ref{fig:mask_ablation_study} demonstrates the improvement in mask generation quality achieved with the RRS module. To evaluate the effectiveness and interaction of each module to the overall performance of the network, we conducted an ablation study on the LOL dataset by systematically removing each module (RRS, TBC, and ACC) and analyzing the impact on performance.

The results are shown in Table~\ref{tab:Ablation}, which shows the contribution of each module to the overall network performance in the form of PSNR. In Case 1, where no modules are included, the PSNR is 21.44, serving as the baseline.
In Case 2, incorporating only the RRS module improves the PSNR to 23.82, highlighting its role in enhancing semantic-level target localization.
In Case 3, adding TBC and RRS modules further improved the PSNR to 22.31, demonstrating its effectiveness in achieving quantitative brightness adjustment. Finally, in Case 4, the highest PSNR (25.78) is achieved due to integrating three modules including the ACC module. This shows that the combination of these interactions significantly enhances the network's context-aware, semantic-level light enhancement capabilities. In summary, this section highlights the complementary contributions of these modules and their importance in achieving good overall performance.

\begin{table}[ht]
\caption{Ablation of RRS, TBC and ACC module on LOL dataset in terms of PSNR.}
\begin{center}
\begin{tabular}{ccccc}
\hline
Case & RRS          & TBC          & ACC          & LOL (PSNR) \\ \hline
1    &              & $\checkmark$ & $\checkmark$ & 21.44      \\
2    & $\checkmark$ &              & $\checkmark$ & 23.82      \\
3    & $\checkmark$ & $\checkmark$ &              & 22.31      \\
4    & $\checkmark$ & $\checkmark$ & $\checkmark$ & 25.78      \\ \hline
\end{tabular}
\label{tab:Ablation}
\end{center}
\end{table}

\subsection{Flexible lighting adjustment by using prompts in an open-world scenario.}
\begin{figure*}
\centering
\includegraphics[width=0.9\textwidth,keepaspectratio]{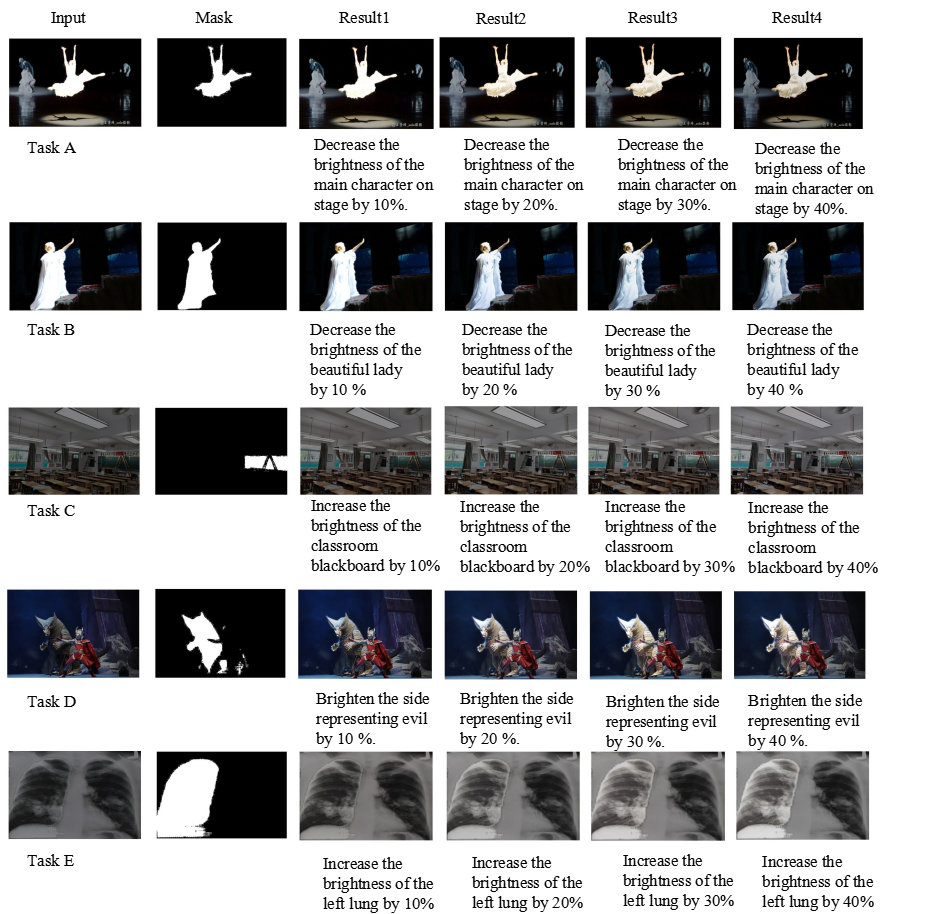}
\caption{Results from different tasks in an open-world scenario. In the case of Tasks A and B, the goal was to decrease the brightness, whereas in the case of some others, the goal was to increase the brightness, covering a wide range of scenes including stage performances, everyday settings, and medical images. Specifically, the tasks are based on increasing or decreasing the brightness of a masked object (the main character, the lady, the blackboard, the side representing evil, and the left lung) by a percentage ranging from 10\% to 40\%.}
\label{fig:local}
\end{figure*}

Leveraging the advanced capabilities of LLM, RRS, TBC and Diffusion, we are able to regulate the overall image brightness to a defined level while making precise adjustments to targeted areas, all through the use of natural language.In Figure~\ref{fig:local}, five distinct tasks in an open-world scenario are illustrated, each focusing on altering the brightness of a particular region in an image. For Task A, the brightness of the main character on stage is progressively decreased by 10\% to 40\%. Similarly, in Task B, the brightness of the lady in the image is reduced by the same increments. Task C focuses on increasing the brightness of a classroom blackboard, while Task D brightens the side representing evil in a fantasy scene. Finally, in Task E, the brightness of the left lung in a medical image is enhanced by increasing steps. The results clearly demonstrate the method's precision in localized brightness adjustment, even with complex prompts and in an open-world scenario.
By interpreting these commands correctly, the system adjusts the brightness of selected areas or objects. In this way, both targeted and background lighting can be adjusted simultaneously.Consequently, the system is capable of performing complex lighting tasks entirely using natural language descriptions. Not only does this demonstrate the flexibility of incorporating language models into image processing, but it also illustrates the potential for dynamic, real-world lighting control applications.    
\subsection{Low-Light Face Detection}
\begin{figure*}[!htbp]
\centering
\resizebox{0.9\textwidth}{!}{%
 \includegraphics{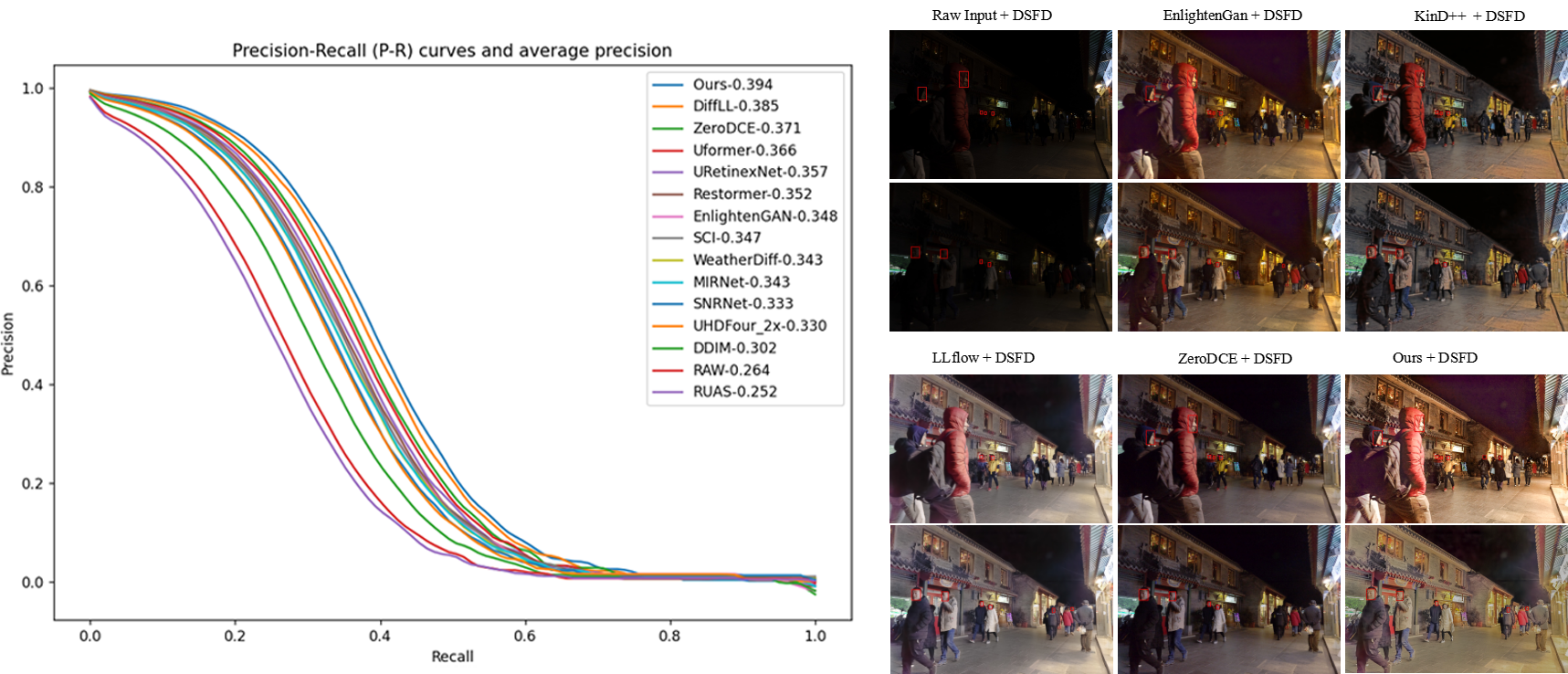}
}
\caption{Face detection performance in low-light conditions is shown across different methods. The left figure displays Precision-Recall (PR) curves, where the "Ours + DSFD" achieves the highest precision. The "Ours + DSFD" delivers the clearest and most accurate results compared to raw input, EnlightenGAN, KinD++, LLflow, and ZeroDCE, in right figure.}
\label{fig:darkface_PR}
\end{figure*}

In this section, Figure ~\ref{fig:darkface_PR} illustrates the performance of various low-light image enhancement methods when applied as preprocessing for face detection in low-light environments.
To begin, the study utilized the DARK FACE dataset ~\cite{yang2020advancing}, which was then followed by a comparison to the widely recognized face detector, DSFD ~\cite{li2018dsfd}.
As shown in Figure \ref{fig:darkface_PR}, the Precision-Recall (PR) curves demonstrate the superior performance of the "Ours + DSFD" method, which achieved the highest average precision (AP) of 0.394. In contrast, other methods, such as ZeroDCE, Uformer, and EnlightenGAN, performed lower on the PR curve.
Additionally, as seen in the figure to the right, the DSFD combined with different enhancement methods is used to detect faces in challenging low-light environments.
Overall, the "Ours + DSFD" approach consistently produces more accurate and clearer face detection results than alternatives such as KinD++, LLflow, and ZeroDCE. Consequently, our method successfully enhances visibility in low-light images, thereby allowing the DSFD detector to operate more effectively in these conditions.

\section{Conclusion}
This work introduces a novel task and a diffusion-based approach for enhancing low-light images using natural language to improve both semantic and quantitative aspects of image enhancement. Our framework integrates advanced modules such as the Retinex-based Reasoning Segment (RRS) for precise target localization, the Text-based Brightness Controllable (TBC) module for natural language-driven brightness adjustment, and the Adaptive Contextual Compensation (ACC) module for seamlessly integrating external conditions across domains. Experimental results demonstrate that our method outperforms state-of-the-art approaches on the LOL and MIT-Adobe FiveK datasets in terms of PSNR, SSIM, and LPIPS metrics, while also showcasing its semantic-level controllability for light enhancement using complex text-driven prompts on open-world datasets. However, the effectiveness of the proposed framework heavily depends on the Large Language Model’s (LLM) ability to accurately interpret and convert natural language instructions into executable quantitative conditions, with language ambiguities posing challenges in target localization and brightness estimation. While this project does not involve color correction, future work will explore how natural language can generate aesthetically pleasing images.

\section{Acknowledgment}
This work is partially supported by Shenzhen Fundamental Research (General Program)(WDZC20231129163533001) and Hunan Provincial
Transportation Science and Technology Project (No.202403).


\bibliographystyle{elsarticle-num}
\bibliography{references1}

\end{document}